# Chest X-ray Classification using Deep Convolution Models on Low-resolution images with Uncertain Labels

Snigdha Agarwal[1], Neelam Sinha[2]

[1]Department of Networking and Communication, International Institute of Information Technology, Bangalore, India
[2]Centre for Brain Research, Indian Institute of Science, Bangalore, India

*Abstract*—Deep Convolutional Neural Networks have consistently proven to achieve state-of-the-art results on a lot of imaging tasks over the past years' majority of which comprise of high-quality data. However, it is important to work on low-resolution images since it could be a cheaper alternative for remote healthcare access where the primary need of automated pathology identification models occurs. Medical diagnosis using low-resolution images is challenging since critical details may not be easily identifiable. In this paper, we report classification results by experimenting on different input image sizes of Chest X-rays to deep CNN models and discuss the feasibility of classification on varying image sizes. We also leverage the noisy labels in the dataset by proposing a Randomized Flipping of labels technique. We use an ensemble of Multi-label classification models on frontal and lateral studies. Our models are trained on 5 out of the 14 chest pathologies of the publicly available CheXpert dataset. We incorporate techniques such as augmentation, regularization for model improvement and use class activation maps to visualize the neural network's decision making. Comparison with classification results on data from 200 subjects, obtained on the corresponding high-resolution images, reported in the original CheXpert paper, has been presented. For pathologies Cardiomegaly, Consolidation and Edema, we obtain 2-3% higher accuracy with our model architecture.

## I. INTRODUCTION

Electromagnetic radiation(X-ray) radiology exam is one of the most common modalities used for identifying different pathological conditions in people because of its simplistic technique and faster diagnosis. It is one of the first tests' done on patients to arrive at an initial diagnosis for a lot of different diseases and is used for different parts of body like teeth, bones, chest etc. [1]. For this paper, we work on a chest x-rays dataset for pathology classification. In chest x-rays, it's not always easy to distinguish between different pathologies even by trained radiologists because of the similarity between the thorax diseases. Hence, a lot of research is in progress to leverage Deep Learning for automated chest pathology classification [2]. Since, it's such a common modality in radiology having a lot of data is ensured which helps us in building more robust deep learning models [3]. Such a model would also prove beneficial in locations where there is a lack of diagnostic expertise and could assist a low experienced radiologist in better decision making [4]. A medical application like this running in remote areas with constraints on investments on high-end infrastructures such as reduced available GPU memory and low network bandwidth may not be feasible. We trained our networks on one of the largest Chest Xrays dataset currently available shared by Stanford Radiology, here after referred to as 'CheXpert'.[5] and evaluated it on a validation set containing 200 studies, which were manually annotated by 3 board-certified radiologists. The target of our research is two-fold. Firstly, we explore network architectures that use smaller images for faster inferencing but do not compromise on model accuracy. We try to achieve better or at par results to the original CheXpert research and secondly, to incorporate the studies labelled as uncertain into the model for better accuracy.

### A. Dataset Description

CheXpert is a large public dataset for chest radiograph interpretation, consisting of 224,316 chest radiographs of 65,240 patients labeled for the presence of 14 observations as positive, negative, or uncertain. A set of 200 patients' records are kept aside as test set. The CheXpert paper presents its results on 5 pathologies out of these 14 which are more clinically relevant than the others. More than one pathology is also possible to exist in the images. The 5 pathologies are:

- Atelectasis
- Cardiomegaly
- Consolidation
- Edema
- Pleural Effusion

Our work also comprises of the same 5 pathologies and we report our results in contrast to the original paper. The dataset comprises of frontal views for all patients and lateral views for few patients. The distribution of these views is given in Fig. 1.

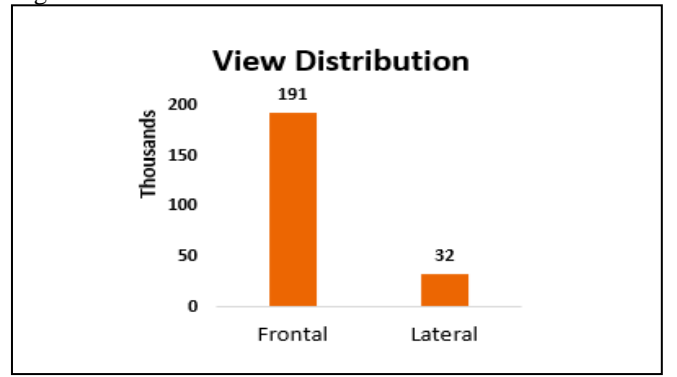

Fig. 1. Frontal and Lateral view image distribution in the CheXpert dataset

The dataset contains uncertainty labels for the existence of pathologies in the x-ray. The uncertainty of a pathology is estimated by an NLP labeler introduced by the paper which automatically identifies the presence of pathologies out of the 14 pathologies by processing the radiologists' notes for patient studies. If the labeler is uncertain about the presence of any pathology, it is given an uncertain label. This identification is used to form the ground truths for the x-ray images. The distribution of positive, negative and uncertain labels in the 5 pathologies is given in Fig. 2. The original CheXpert paper talks about handling uncertainty labels in 5 ways. The approaches to handle uncertainty labels are:

*1) U-Ignore* - This approach ignores all the records which have uncertainty in any pathology, thus restricting to only the rightly classified images

*2) U-Zeros* - This approach considers all records with uncertain pathologies as negative cases (i.e. not present)

*3) U-Ones* - This approach considers all records with uncertain pathologies as positive cases(i.e. present)

*4) U-SelfTrained* - This approach keeps aside all the cases which have uncertainty classes and using U-Ignore model tries to predict the missing classes

*5) U-Multiclass* - This approach keeps all the three classes as separate classes and then classify using individual model [6]

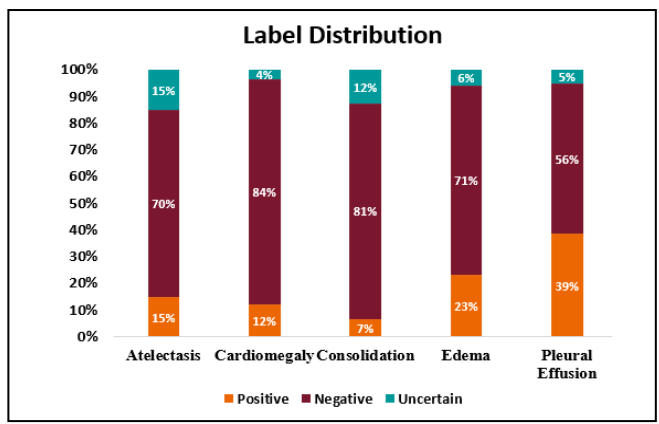

Fig. 2. Positive, negative and uncertain label distribution in the CheXpert dataset for the 5 pathological classes

## II. RELATED WORK

1. Over the years, few large Chest X-ray databases have been made public by the research community. Since, Chest X-rays can be associated with multiple pathologies at the same time, a lot of work has been done in Multi-Label Classification [7] or like some research communities call it Multi-Task Classification [8]. Due to the importance of X-rays in medical imaging, a variety of approaches have been proposed. CheXnet, which was a classification on 14 pathologies was proposed by the CheXpert authors on the ChestX-ray14 dataset [9,10] using the Dense-Net121. Chen et al. [11] proposed to use the conditional training strategy to exploit the hierarchy of lung abnormalities in the PLCO dataset [12]. Baltruschat et al. [13] presented a comparison on Multi-Task Classification using the ChestX-ray14 dataset with different deep learning architectures of ResNet [14] and DenseNet [15]. Allaouzi et al. [16] proposed an architecture for the CheXpert problem which extracts feature vectors of each image from a deep learning model and test on different machine learning classification strategies on top of it. Their best reported accuracy on the data is 81%. In [28], the authors propose an Attention induced pure convolutional neural network architecture for classification of intracranial hemorrhages with an average sensitivity of 93% and average precision of 92%.

## III. PROPOSED METHOD

For building classification models for the 5 pathologies we have used Convolutional Neural Networks (CNNs) [17] which are one class of Deep Learning models for image data. Since, the dataset comprises of studies with multiple pathologies at the same time, we have approached the problem as Multi-Label [7] Classification. We feed images for all the 5 pathologies together to a single classification model with each image having one or more pathologies amongst the 5.

### A. Handling the Uncertainty Labels

The uncertainty labels are quite prevalent in few of the pathologies such as Consolidation and Atelectasis as seen in Figure 2. Ignoring these uncertain labelled studies (U-Ignore approach) completely for the training would lead to a poor model which is unaccounted for a lot of borderline cases. Also, marking them all as positive or negative (U-Ones and U-Zeros approach) might lead to a model that has learnt a lot of wrong features. To rectify this problem, we have used a novel solution of Randomized flipping of the uncertain labels to positive and negative. This random selection would help us mitigate a risk of inaccurate learning through U-Zeros or U-Ones approach and no learning from the U-Ignore approach. The labeling of the uncertain studies follows the below formula,

$$y^i = \text{random } (0,1,0.1) \qquad (2)$$

$$y = \text{if } y^i \geq 0.5, 1 \text{ else } 0 \qquad (3)$$

### B. Model Architecture

As an initial step in building the CNN for the model, we started by stacking convolutional and max pooling layers as Conv→Conv→Pool order. This was to see if we can achieve any learning by making shallow models or Feed forward networks [18]. This basic block is stacked one over the other to 10 such blocks. For regularization of the model, we added dropout layer after every 3rd convolution block. With such a kind of architecture we were getting an accuracy of about 35%.We found that our model needs to go deeper for learning more features for this data as is pointed out by numerous researches before on the benefits of deep neural networks over shallow ones [19]. For this, instead of stacking more blocks of our simple convolution, we decided to work with the named CNN architectures as a lot of work on the robustness of multiple CNN architectures is already established [20]. For this paper, we have worked with the Densely Connected Convolutional Networks (DenseNet) architecture. We follow the same architecture as presented by the original DenseNet paper[15]. We go with a depth of 121 layers for this architecture. The activation of the last layer is changed to sigmoid as opposed to softmax because of the multiple labels. The sigmoid activation values are then treated as individual probabilities of each class [21]. The Dense Convolutional Network (DenseNet), connects each layer to every other layer in a feed-forward fashion. Whereas traditional convolutional networks with L layers have L connections—one between each layer and its subsequent layer—Densenet has L(L+1)/ 2 direct connections. For each layer, the feature-maps of all preceding layers are used as inputs, and its own feature-maps are used as inputs into all subsequent layers. Some of the advantages of using such an architecture are,

- alleviates the vanishing-gradient problem
- strengthens feature propagation
- encourages feature reuse

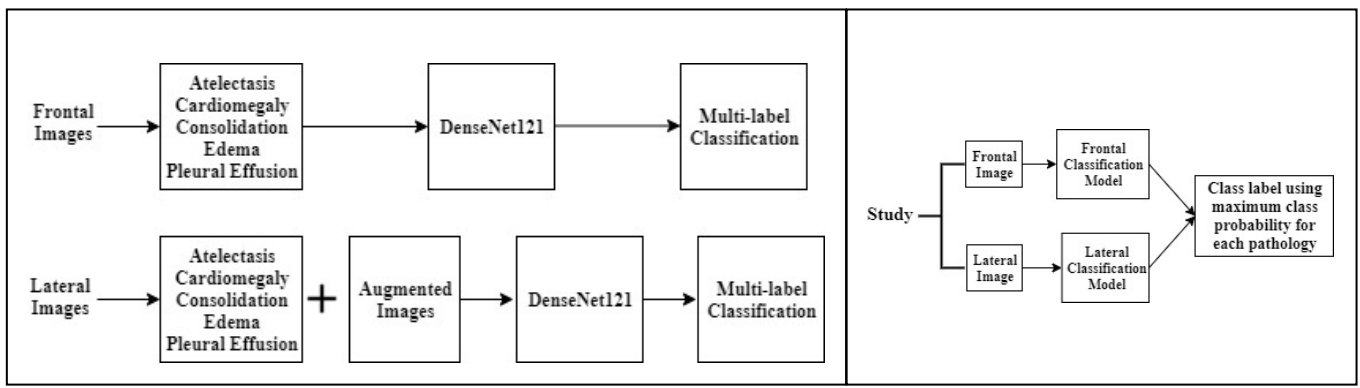

Fig. 3. The left part shows the training architecture used for each of the studies. The models are trained separately for frontal and lateral images. The right part of the image shows the inferencing of each study in the validation dataset. If study has both frontal and lateral images, maximum class probability is used for final class label.

- reduces the number of parameters.

We train two different Multi-label classification models, for the frontal and lateral views. The model is run multiple times and the weights of the 5 best checkpoints are ensembled into one model. The trained models are then used for inferencing on the validation dataset for the performance analysis. If one patient study has both frontal and lateral images, the class label is decided based on maximum class probability from each of the images. The model workflow for training and inferencing is shown in Fig. 3.

Using this architecture, we experiment with three different input sizes to the neural network architecture, 224x224, 197x197 and 139x139. We try to assess the minimum image quality required for such an x-ray classification task without compromising on the classification results. We discuss the results in Section V.

*C. Network Enhancements*

We fix the architecture for our network and work towards enhancements on our network to improve on the model robustness and accuracy. We use grid search and fine scanning strategy for hyperparameter tuning. We discuss some of the enhancements and processes we did during model building and tuning here,

*1) Resizing Images and Preprocessing:* The images are preprocessed per the selected input sizes as discussed in Section III, using a Lambda layer after the initial input layer. We have also used a Caffe style preprocessing for the images, which is basically mean subtraction and zero centering using mean pixel.

*2) Working with imbalanced classes:* Since the number of cases in each of the pathologies are different, we would be working with imbalanced classes. Imbalanced classes lead to the model being bias towards the majority class and might give false impressions of accuracy [22]. To solve for this, we have balanced the class weights by giving an extra parameter during defining the network. The class weights are calculated by the below formula,

$\omega$ = Number of studies in class/Total Studies in all class    (1)

*3) Data Augmentation:* To improve the models further in making it train on multiple variations of data, we have fixed on augmentation parameters after going through different values for each parameter. We have experimented with different parameters by seeing the images to how much augmentation wouldn't distort the complete structure of the images. For the frontal images, since the data is sufficient we directly apply augmentation on the studies itself. For lateral images, we are increasing the data size by creating 3 more studies per each study. This helps us in training the deep learning model on sufficient data. We restrict the augmentation strategy from any of the horizontal or vertical flips. The height and width shift range are restricted to 10%. Rotation range has been kept between 0 to 3. A slight zoom range is also applied to each of the images.

*4) Regularization:* The original CheXpert paper uses the high-resolution images for its experiments. They run their model to up to 3 epochs for each experiment. We have not restricted our model to this condition and leverage the predictive power up to the lowest loss value possible by running through multiple epochs. To come out of conditions of overfitting when running for higher epochs we have used regularization in the model by adding l2 regularizer (L2

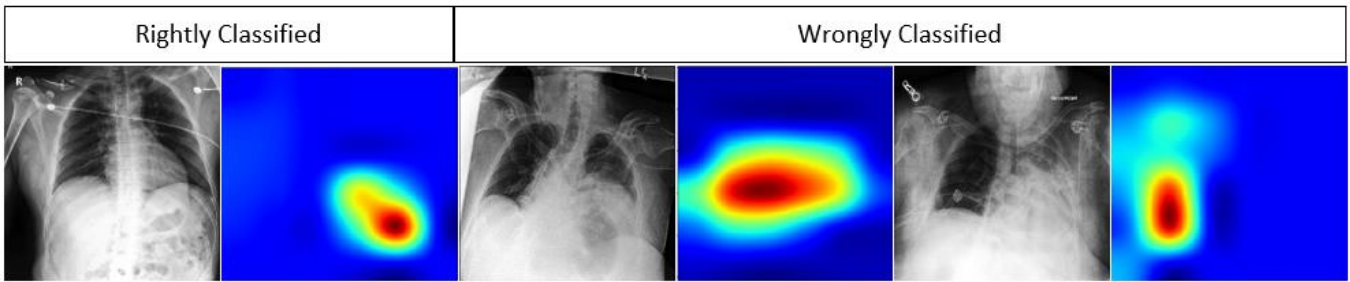

Fig. 4. Class activation maps for the last layer of the model for Cardiomegaly pathology. The left part of the image shows activations for rightly classified images and right part shows activations for some mis-classified image

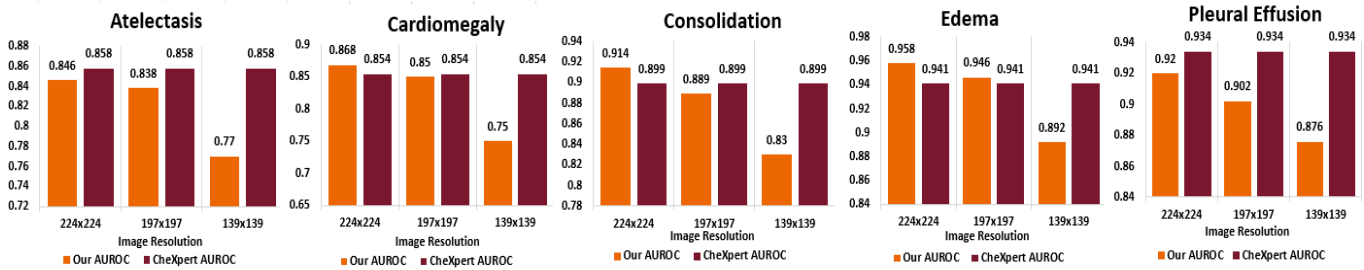

Fig. 5. Comparison charts for each of the 5 pathologies for our models with CheXpert results. We can see that our models are at par or slightly better than some of the CheXpert results for few pathologies even for lower resolution images.

norm) [23] in our model definition.

*5) Learning Rate Decay:* While running for higher epochs, we decrease the learning rate by a very small percentage of 0.1 of the initial learning rates at plateaus to prevent the model from getting stuck at local minima after a patience period of 3 epochs. We also stop the run early if there is no change in loss for about 10 consecutive epochs.

*6) Hyperparameter optimization*: For deciding on the best hyperparameters for the model, we have used a "coarse search strategy" where we try to identify the best hyperparameters on a smaller dataset using a grid search technique[24] to identify the bounds in which our best parameter will lie and then running "finer search" for the specific parameters on the whole dataset.

## IV. VISUALIZATION

Further to this, we have also visualized the learning of our model using class activation maps [25]. This process helps us in identifying the reasons for false positives and false negatives in the model. We visualize the concentration of model's important features for each class and see if the model has the capability of reading the right features for distinction. We find out that for wrongly classified images the concentration of features more attentive to the network are directed towards different locations with concentrated white pixels, while the model for the correct classification concentrates more on specific regions of localized pathological conditions. This visualization proves that the model has the ability of reading the right pathological features for classification. Our visualization results are presented in Fig. 5.

## V. RESULTS

We compare our results with the CheXpert baseline model considering their best uncertainty approach using the AUROC metric as is defined in [26]. We show that by going from image size of 224x224 to 197x197 doesn't make a significant change in accuracy while going from 197x197 to 139x139 makes a statistically significant difference between the accuracies with the baseline. It is to be noted that significance is based on t-test [27] among multiple runs of each model. The results are summarized in Fig. 5. We also show that by using our Randomized Flipping technique in dealing with the uncertainly labelled studies we are able to capture the variance in the images as can be told by the AUROC without losing any relevant information as compared to ignoring all uncertainties(U-Ignore approach) or making the model read contrasting details by considering all as positive(U-Ones approach) or negative(U-Zeros approach).

## VI. CONCLUSION

We successfully present a neural network architecture considering uncertainty labels with a novel selection technique using a relatively low-resolution input and using an ensemble of just two models.

REFERENCES

[1] Howell, Joel D. "EARLY CLINICAL USE OF THE X-RAY." Transactions of the American Clinical and Climatological Association vol. 127 (2016): 341-349.

[2] Baltruschat, I.M., Nickisch, H., Grass, M. et al. Comparison of Deep Learning Approaches for Multi-Label Chest X-Ray Classification. Sci Rep 9, 6381 (2019) doi:10.1038/s41598-019-42294-8.

[3] X. Chen and X. Lin, "Big Data Deep Learning: Challenges and Perspectives," in *IEEE Access*, vol. 2, pp. 514-525, 2014.

[4] Gabriel Chartrand, Phillip M. Cheng, Eugene Vorontsov, Michal Drozdzal, Simon Turcotte, Christopher J. Pal, Samuel Kadoury, and An Tang, "Deep Learning: A Primer for Radiologists", in RadioGraphics 2017 37:7, 2113-2131.

[5] J. Irvin, P. Rajpurkar, M. Ko, Y. Yu, S. Ciurea-Ilcus, C. Chute, H. Marklund, B. Haghgoo, R. L. Ball, K. Shpanskaya, J. Seekins, D. A. Mong, S. S. Halabi, J. K. Sandberg, R. Jones, D. B. Larson, C. P.

Langlotz, B. N. Patel, M. P. Lungren, A. Y. Ng, CheXpert: A large chest radiograph dataset with uncertainty labels and expert comparison, in: AAAI, 2019.

[6] Erin Allwein, Robert Shapire, and Yoram Singer. Reducing multiclass to binary: A unifying approach for margin classifiers. Journal of Machine Learning Research, pages 113–141, 2000.

[7] M.-L. Zhang, Z.-H. Zhou, A review on multi-label learning algorithms, IEEE Transactions on Knowledge and Data Engineering 26 (8) (2013) 1819–1837.

[8] G. Tsoumakas, I. Katakis, Multi-label classification: An overview, International Journal of Data Warehousing and Mining 3 (3) (2007) 1–13.

[9] P. Rajpurkar, J. Irvin, K. Zhu, B. Yang, H. Mehta, T. Duan, D. Ding, A. Bagul, C. Langlotz, K. Shpanskaya, et al., ChexNet: Radiologist-level pneumonia detection on chest X-rays with deep learning, arXiv preprint arXiv:1711.05225.

[10] P. Rajpurkar, J. Irvin, R. L. Ball, K. Zhu, B. Yang, H. Mehta, T. Duan, D. Ding, A. Bagul, C. P. Langlotz, et al., Deep learning for chest radiograph diagnosis: A retrospective comparison of the CheXNeXt algorithm to practicing radiologists, PLoS Medicine 15 (11) (2018) e1002686. doi:https://doi.org/10.1371/journal.pmed.1002686.

[11] H. Chen, S. Miao, D. Xu, G. D. Hager, A. P. Harrison, Deep hierarchical multi-label classification of chest X-ray images, in: MIDL, 2019, pp. 109–120.

[12] J. K. Gohagan, P. C. Prorok, R. B. Hayes, B.-S. Kramer, The prostate, lung, colorectal and ovarian (plco) cancer screening trial of the national cancer institute: History, organization, and status, Controlled Clinical Trials 21 (6, Supplement 1) (2000) 251S – 272S. doi:https://doi.org/10.1016/ S0197-2456(00)00097-0.

[13] Baltruschat, I.M., Nickisch, H., Grass, M. et al. Comparison of Deep Learning Approaches for Multi-Label Chest X-Ray Classification. Sci Rep 9, 6381 (2019) doi:10.1038/s41598-019-42294-8.

[14] Deep Residual Learning for Image Recognition, Kaiming He, Xiangyu Zhang, Shaoqing Ren, Jian Sun, 2015.

[15] Densely Connected Convolutional Networks, Gao Huang, Zhuang Liu, Lausrens van der Maaten, Kilian Q. Weinberger, 2016.

[16] Allaouzi, Imane & Ahmed, M.. (2019). A novel approach for multi-label Chest X-Ray classification of common Thorax Diseases. IEEE Access. PP. 1-1. 10.1109/ACCESS.2019.2916849.

[17] Xu, L. & Ren, Jimmy & Liu, C. & Jia, J.. (2014). Deep convolutional neural network for image deconvolution. Advances in Neural Information Processing Systems. 2. 1790-1798.

[18] GOODFELLOW, I., BENGIO, Y., & COURVILLE, A. (2016). Deep learning. Chapter 6.

[19] Winkler, David. (2018). Deep and Shallow Neural Networks: Basic Concepts and Methods. 10.1002/9783527816880.ch11_03.

[20] The History Began from AlexNet: A Comprehensive Survey on Deep Learning Approaches, Md Zahangir Alom, Tarek M. Taha, Chris Yakopcic, Stefan Westberg, 2018

[21] Maxwell, A., Li, R., Yang, B. et al. Deep learning architectures for multi-label classification of intelligent health risk prediction. BMC Bioinformatics 18, 523 (2017).

[22] Kotsiantis, Sotiris & Kanellopoulos, D. & Pintelas, P.. (2005). Handling imbalanced datasets: A review. GESTS International Transactions on Computer Science and Engineering. 30. 25-36.

[23] Cortes, C., Mohri, M., & Rostamizadeh, A. (2012). L2 regularization for learning kernels. arXiv preprint arXiv:1205.2653.

[24] Random Search for Hyper-Parameter Optimization, James Bergstra, Yoshua Bengio, 2012.

[25] Grad-CAM: Visual Explanations from Deep Networks via Gradient-based Localization, Ramprasaath R. Selvaraju, 2017.

[26] Kim, Tae Kyun. "T test as a parametric statistic." Korean journal of anesthesiology vol. 68,6 (2015): 540-6. doi:10.4097/kjae.2015.68.6.540

[27] Hajian-Tilaki, Karimollah. "Receiver Operating Characteristic (ROC) Curve Analysis for Medical Diagnostic Test Evaluation." Caspian journal of internal medicine vol. 4,2 (2013): 627-35.

[28] S. Agarwal, C. S. Pradeep and N. Sinha, "Temporal Surgical Gesture Segmentation and Classification in Multi-gesture Robotic Surgery using Fine-tuned features and Calibrated MS-TCN," *2022 IEEE International Conference on Signal Processing and Communications (SPCOM)*, Bangalore, India, 2022, pp. 1-5, doi: 10.1109/SPCOM55316.2022.9840779.